# AGENT-BASED ECOLOGICAL MODEL CALIBRATION – ON THE EDGE OF A NEW APPROACH

António Pereira[1], Pedro Duarte[1], Luís Paulo Reis[2]
apereira@ufp.pt, pduarte@ufp.pt, lpreis@fe.up.pt
[1]CEMAS –Center for Modeling and Analysis of Environmental Systems,
Faculty of Science and Technology, University Fernando Pessoa, Portugal
[2]LIACC – Artificial Intelligence and Computer Science Lab.,
Faculty of Engineering, University of Porto, Portugal

**Abstract** – In every mathematical model, parameters regulate the behaviour of equations describing temporal and spatial changes of model state variables and their interactions. Generally, there is some uncertainty associated with each parameter. Model calibration is performed by comparing observed with predicted data and it is a crucial phase in the modelling process. It's an iterative and interactive task in which, after each simulation, the "modeller" analyses the results and performs changes on one or more equation's parameters trying to tune the model. This "tuning" procedure is a hard and "tedious" work requiring a good understanding of the effects of different parameters over the available variables. Automatic calibration procedures, based on systematic and exhaustive generation of parameter vectors and using several convergence methods, are available but they require a large number of model runs and are, therefore, not applicable to very complex ecosystem models demanding large computational times. A possible alternative may be to develop a self-learning tool that simulates the learning process of the modeller about the simulated system. The purpose of this paper is to present a new approach to ecological model calibration – an agent-based software. This agent works on three stages: (i) It builds a matrix that synthesizes the inter-variable relationships; (ii) It analyses the steady-state sensitivity of different variables to different parameters; (iii) It runs the model iteratively and measures model lack of fit, adequacy and reliability. Stage (iii) continues until some convergence criteria are attained. At each iteration, the agent "knows" from (i) and (ii), which parameters are most likely to produce the "desired" shift on predicted results.

## I. INTRODUCTION

Many mathematical models used in the fields of ecology, economics and environmental science are based on a body of knowledge formed with not generally accepted theories, debatable or controversial hypothesis, questionable simplifications and a bundle of implicit or ambiguous assumptions, i.e., based on an imperfect understanding of the dynamics of the object systems. This leads to highly uncertain model results because of the uncertainty associated with model parameters and inputs and, sometimes, the uncertainty in model structure [1].

When an ecological model is built, those uncertainties are intrinsic to the model and the major problem is to quantify the quality of the simulations in order to recognize if a modification of the concepts, laws simulating the processes or model parameters would improve it [2]. If the concepts and laws of the simulated processes are well established, attention must be directed to deciding parameter values. Calibration of these parameters, i.e., defining appropriate values for each parameter in the simulation in order to approximate simulation results to reality, is a task of major importance.

Several procedures for automatic calibration and validation are available in the literature, like the Controlled Random Search (CRS) method [1][3] or linear regression techniques [2]. However, these procedures do not capture the complexity of human reasoning in the calibration process. They are based on the systematic and exhaustive generation of parameter vectors and require a large number of model runs, demanding heavy computationally search operations. In addition, when the model is very complex, those procedures demand large computational time.

The traditional calibration is oriented, i.e., the "modeller" analyses the results and, in face of his knowledge about the behaviour of different mathematical relationships, some common sense reasoning is used to choose new values for each parameter. The systematic approach described in [4] argues that the ultimate use of the model should be explicitly acknowledged in the calibration process. These procedures raise the question: "Is it possible to implement that common sense reasoning in an automatic calibration system when the model is very complex?" Being able to answer this question raises an even more challengeable one: "Is it possible to implement a generic automatic calibration system that learns for itself and is self-adaptable to any model?"

This paper introduces a new approach to answer these two questions: an agent-based calibration software. The architecture for the calibration system described herein is based on the "intelligent agents" approach [5][6][7][8]. An agent may be defined as a self-contained software program, specialized in achieving a set of goals, by autonomously performing tasks on behalf of users or other agents. Agents are particularly useful in complex, inaccessible and dynamic environments as ecosystems or other biological systems.

The approach presented in this study is based on a software agent, called Calibration Agent that builds the inter-variable relationships and analyses variable's sensitivity to different parameter changes. The Calibration Agent executes the simulation model iteratively, measuring the lack of fit, adequacy and reliability [1][3] at each round, until some predefined convergence criteria is attained. At each simulation iteration, the agent changes values of selected parameters trying to minimize the lack of fit of the results achieved to real data, thus improving the reliability of the model without reducing the adequacy too much [1][3].

This paper is organized as follows. Section II describes the type of ecological modelling problems under analysis in this study and refers some examples. The next section briefly describes the simulation system built under this project, EcoDyn application and its main features. The calibration agent approach is described in section IV. The paper concludes with project state and pointers to future work.

## II. ECOLOGICAL MODELLING

Ecological models are simplified views of nature used to solve scientific or management problems. These models only contain the characteristic features that are essential in the context of the problem to be solved or described. Ecological models may be considered a synthesis of what is known about the ecosystem with reference to the considered problem, as opposed to a statistical analysis - a model is able to translate our knowledge about the system processes, formulated in mathematical equations, and component relationships and not only relationships between data [9].

Ecological models include physical, chemical and biological processes to describe the main features of the ecosystem in analysis. The description of the physical, chemical and biological processes and the system structure do not account for all the details. Carefully designed models, which include important processes and components, still omit details that aren't important to the problem under consideration – many irrelevant details would cloud the main objectives of a model. However, these omitted details might have a strong influence on the predicted output those models produce [9].

In ecological models of aquatic systems, physical processes include flow and circulation patterns, mixing and dispersion of mass and heat, water temperature, settling of planktonic organisms and suspended matter, insulation and light penetration. The simulation of these processes is very important for setting up a good model of the whole ecosystem and detailed descriptions of them are available and widely accepted by modellers. One of the most important compromises is to find the optimal time and space scales of the model. Spatial grids acceptable for physical and chemical processes (10 to 100 metres) are very detailed for biological processes, and similarly, minutes or hours are good time scales for physical and chemical processes, but hours, days and months may be appropriate time scales for biotic components of an ecosystem [9]. The space division must account for variations in horizontal and vertical dimensions. The simplest geometric representation is the zero-dimensional (0D) model, which simulates the system as a point and all changes are only time dependent. One-dimensional (1D) representation models assume that the system is characterized by a prevailing one-directional flow (horizontal or vertical) and the properties of the system vary along that direction and time. When the system is large enough to present sensible variation of the properties, vertical and/or horizontal division is required and two or three-dimensional (2D or 3D) representations are more common. Models of deep large lakes, deep bays or large river estuaries are examples of these representations.

Unlike the chemical and physical parameters that are almost known as exact values, it is rather unusual to know exact values for most biological parameters. Almost all literature about this subject presents biological parameters as approximate values or intervals [9]. Under this context, it is obvious that there is a particular need for parameter estimation methods for most biological parameters. Thus, the need for calibration is therefore "intrinsic" to ecological models [9].

The authors are particularly concerned with coastal lagoons and ecosystems. Located between land and open sea, these ecosystems receive fresh water inputs, rich in organic and mineral nutrients derived from urban, agricultural and industrial effluents and domestic sewage. Furthermore, coastal ecosystems are subject to strong anthropogenic pressures due to tourism and shellfish/fish farming. These factors are responsible for important ecosystem changes characterized by eutrophic conditions, algal blooms, oxygen depletion and hydrogen sulphide production [10]. Examples of ecological models can be found in [7][12][13].

Particularly complex is the ecological modelling developed for Sungo Bay, located in Shandong Province of People's Republic of China [13].

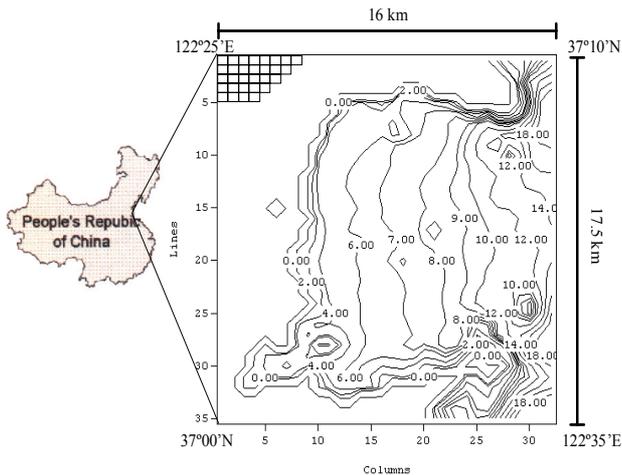

**Figure 1-** Location of Sungo Bay, including model domain and bathimetry(m). Also shown, is a part of the model grid with a spatial resolution of 500m. [13]

The Sungo Bay is modelled as a 2D vertically integrated, coupled hydrodynamic-biogeochemical model. It is based on a finite difference bathymetric staggered grid [14] with 1120 cells (32 columns x 35 lines) and a spatial resolution of 500m (Figure 1). The model time step is 18 seconds. The model has a land and an ocean boundary. It is forced by tidal height at the sea boundary, light intensity, air temperature, wind speed, cloud cover and boundary conditions for some of the simulated state variables and solves the general 2D transport equation[15][16]. The hydrodynamic sub-model solves the speed components, whereas biogeochemical processes such as primary productivity and grazing, as well as physical processes such as sediment deposition and resuspension provide the sources and sinks terms.

The suspended matter dynamics and biogeochemical processes are described as differential equations [13] and are presented in Figure 2. Some of the variables simulated were: water temperature, current speed and direction, dissolved substances, suspended matter, phytoplankton, zooplankton, kelps and bivalve species.

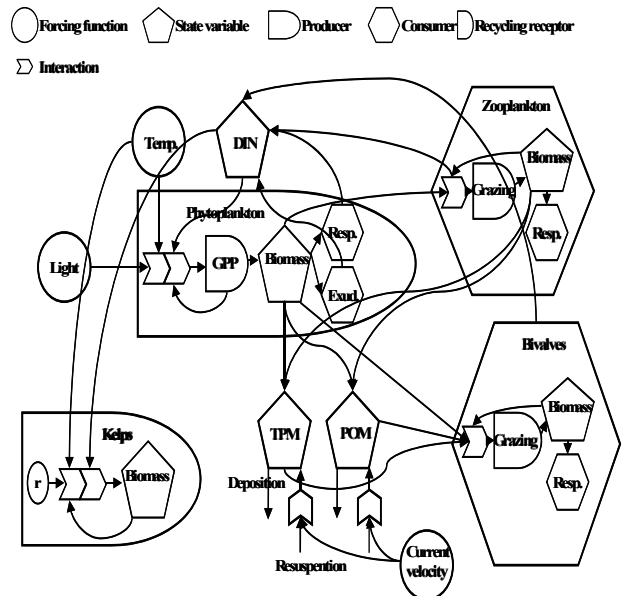

**Figure 2-** Sungo Bay model diagram following the "energy circuit language"[17]

### III. SIMULATION SYSTEM – ECODYN

EcoDyn is an application built to enable physical and biogeochemical simulation of aquatic ecosystems. It's an object oriented program application, built in C++ language, with a shell that manages the graphical user interface (Figure 3), the communications between classes and the output devices where the simulation results are saved. The simulated processes include:
- hydrodynamics of aquatic ecosystems – current speeds, and directions;
- thermodynamics – energy balances between water and atmosphere and water temperature;
- biogeochemical – nutrient and biological species dynamics;
- anthropogenic pressures, such as biomass harvesting.

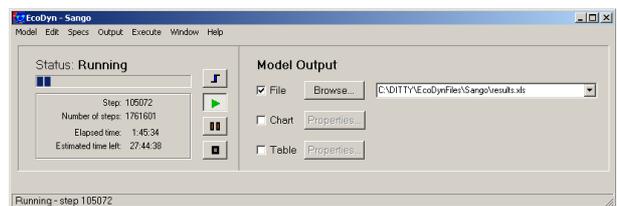

**Figure 3-** EcoDyn shell graphical interface (main window)

The ecosystem characteristic properties are described in a model database: definitions as morphology, geometric representation of the model, dimensions, number of cells, classes, variables, parameter initial values and ranges are present.

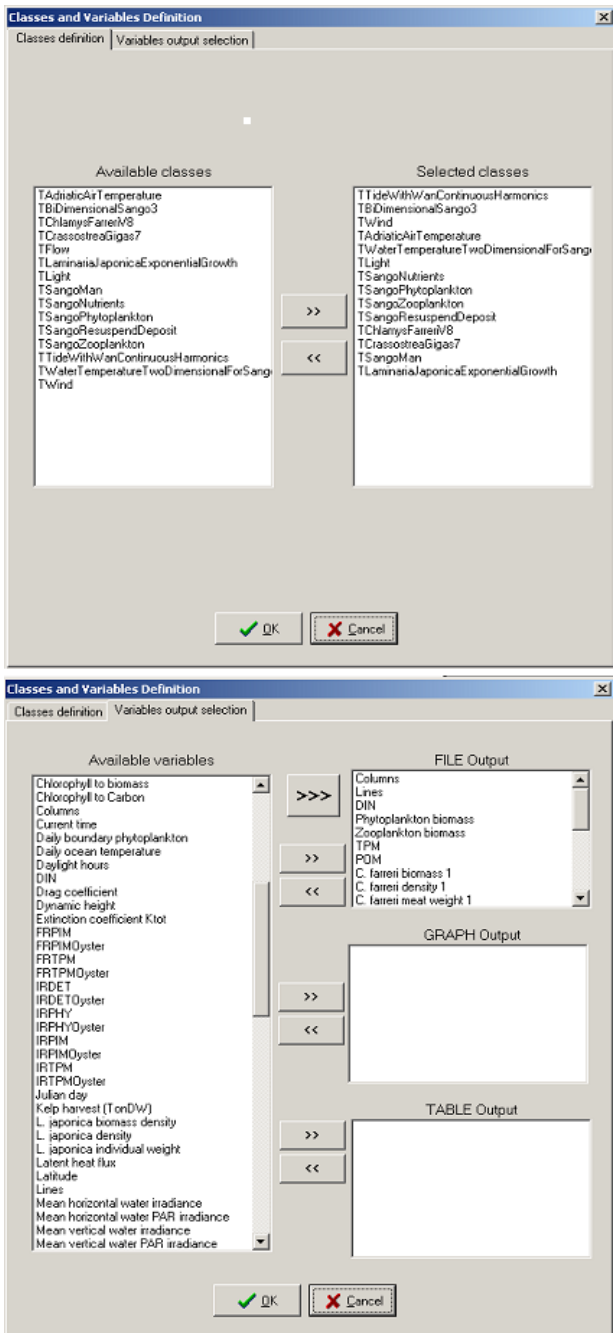

**Figure 4-** Classes and Variables definition dialogs

The user can choose between file, chart or table to store the simulation results. These output formats are compatible with some commercial software products, enabling their posterior treatment.

Different classes simulate different variables and processes, with proper parameters and process equations. Classes can be selected or deselected from shell dialogs (Figure 4) determining its inclusion or exclusion in each simulation of the model.

To provide mutual influence between variables, EcoDyn implements two public communication methods: **Inquiry** and **Update**. Any class may get information about any other class and respective variables using the former method. The inquiring class may be influenced by variables of the inquired one. The later method is used when the invoking class influences variables belonging to the invoked class. All communication between classes occurs through the EcoDyn shell. The invoking and the invoked classes are identified by a name and a code.

This application has an interface module that enables remote control from external/remote applications (typically the Agents). The remote application can do everything the user can (start/stop the EcoDyn application and control the model simulation runs: start, stop, pause, restart and step) and, additionally, can "spy" the simulation activity and change the values of the EcoDyn parameters. When EcoDyn is under the remote control the user interface can be activated only for information. The remote control has precedence over the user control.

## IV. CALIBRATION AGENT

Model calibration is performed by comparing observed with predicted data and is a crucial phase in the modelling process. It's an iterative and interactive task in which, after each simulation, the "modeller" analyses the results and changes one or more equation parameters trying to tune the model. This "tuning" procedure requires a good understanding of the effect of different parameters over different variables.

Evaluation of the result's quality is an easy task with simple algorithms (ex. linear regression between predicted and observed data), the system can classify the results quality in a qualitative or quantitative scale. A more complex problem is the selection of new parameter values to use in the next iteration by the model equations, trying to maximize the model quality of fit.

One way of doing this is to give to the software agent a list with all changeable equation parameters, all possible ranges for those parameters and let it exhaustively search through all available parameter combinations until it finds the optimal one. This is a very intensive computation process due to its uninformed (and thus not intelligent) search through the system's tens or hundreds of equations and parameters. Research on this matter should therefore be focused on devising intelligent search techniques that may be able to use the modeller's knowledge to guide the search.

Knowledge about the behaviour of all system processes, possessed by the "modeller" in the traditional calibration processes, shall be used to guide the selection of the new values for the parameters contained in different mathematical relationships. In the present system, the intelligent agent learns this knowledge in three phases:

- Building matrices that synthesize the interclass and inter-variable relationships;
- Analysing the intra and interclass steady-state sensitivity of different variables to different parameters and among variables;

- Iterative model execution, measuring model lack of fit, adequacy and reliability [1][3] until a convergence criteria is matched.

This methodology gives the Calibration Agent the generality needed to be able to calibrate "any" type of model. The calibration procedure diagram is shown in Figure 5.

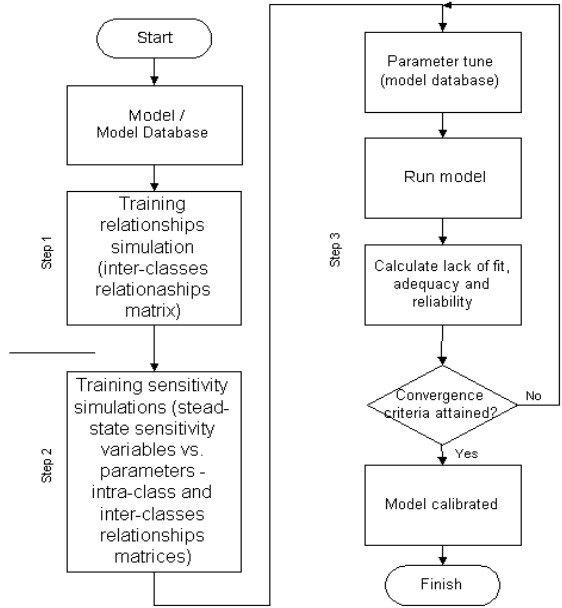

**Figure 5-** Calibration Agent procedure diagram

The first step consists in choosing the model and checking if its database is populated. The Calibration Agent has access to the EcoDyn shell and whenever **Inquiry** and **Update** methods are called it collects the class names, variables and values involved. The agent runs the model for some time just to gather this information ("Training relationships simulation" box). It stores the information in its "knowledge database" as a square matrix, synthesising the relationships between different classes. From the example presented in Table I (model described in [13]) it follows that class TAdriaticAirTemperature influences classes TWaterTemperatureTwoDimensionalForSango and TLight, class TSangoResuspendDeposit influences classes TLight, TSangoPhytoplankton, TSangoNutrients, TChlamysFarreriV8 and TCrassostreaGigas7, class TSangoPhytoplankton influences classes TSangoResuspendDeposit, TSangoNutrients, TSangoZooplankton, TChlamysFarreriV8 and TCrassostreaGigas7, and so on.

The next step is used by the Calibration Agent to perform an exhaustive sensitivity analysis and to synthesize the results obtained in several matrices. First, the intra-class sensitivity is analysed (sensitivity of each variable to each parameter of its own class). Average sensitivities are calculated from parameter ranges defined by the user in the model database. The results of this analysis are stored in one matrix per class like the one exemplified in Table II – class A with nA variables and mA parameters: changes imposed on parameter P1A affects variables V1A and V2A by 0.3 % and –0.6%, respectively.

**Table I-** Square matrix synthesizing relationships between different classes

| | TWaterTemperatureTwoDimensionalForSango | TBiDimensionalSango3 | TAdriaticAirTemperature | TLight | TSangoResuspendDeposit | TWind | TSangoPhytoplankton | TSangoNutrients | TTideWithContinuousVanHarmonics | TLaminariaJaponicaExponentialGrowth | TSangoZooplankton | TChlamysFarreriV8 | TCrassostreaGigas7 |
|---|---|---|---|---|---|---|---|---|---|---|---|---|---|
| TWaterTemperatureTwoDimensionalForSango | - | 1 | 0 | 0 | 0 | 0 | 0 | 0 | 0 | 1 | 0 | 0 | 0 |
| TBiDimensionalSango3 | 1 | - | 0 | 0 | 1 | 0 | 1 | 1 | 0 | 1 | 1 | 1 | 1 |
| TAdriaticAirTemperature | 1 | 0 | - | 1 | 0 | 0 | 0 | 0 | 0 | 0 | 0 | 0 | 0 |
| TLight | 1 | 0 | 0 | - | 0 | 0 | 1 | 0 | 0 | 0 | 0 | 0 | 0 |
| TSangoResuspendDeposit | 0 | 0 | 0 | 1 | - | 0 | 1 | 1 | 0 | 0 | 0 | 1 | 1 |
| TWind | 1 | 0 | 0 | 0 | 0 | - | 0 | 0 | 0 | 0 | 0 | 0 | 0 |
| TSangoPhytoplankton | 0 | 0 | 0 | 0 | 1 | 0 | - | 1 | 0 | 0 | 1 | 1 | 1 |
| TSangoNutrients | 0 | 0 | 0 | 0 | 0 | 0 | 0 | - | 0 | 1 | 0 | 0 | 0 |
| TTideWithContinuousVanHarmonics | 0 | 1 | 0 | 0 | 0 | 0 | 0 | 0 | - | 0 | 0 | 0 | 0 |
| TLaminariaJaponicaExponentialGrowth | 0 | 0 | 0 | 0 | 0 | 0 | 0 | 1 | 0 | - | 0 | 0 | 0 |
| TSangoZooplankton | 0 | 0 | 0 | 0 | 1 | 0 | 0 | 0 | 0 | 0 | - | 0 | 0 |
| TChlamysFarreriV8 | 0 | 0 | 0 | 0 | 1 | 0 | 1 | 0 | 0 | 0 | 0 | - | 0 |
| TCrassostreaGigas7 | 0 | 0 | 0 | 0 | 1 | 0 | 1 | 0 | 0 | 0 | 0 | 0 | - |

**Table II-** Matrix synthesizing intra-class sensitivity coefficients between variables and parameters

| Class A | Parm P1A | Parm P2A | ... | Parm PmA |
|---|---|---|---|---|
| Var V1A | 0.003 | 0 | | 0.34 |
| Var V2A | -0.006 | 0.004 | | 0 |
| ... | | | | |
| Var VnA | 0 | 0 | | -0.0071 |

Secondly, the inter-class sensitivity is analysed (sensitivity of each variable of each class is analysed with respect to all variables of each class by which it is influenced). During this step, the model runs ("Training sensitivity simulation" box) keeping all variables and parameters constant except those directly involved in sensitivity analysis.

Since intra and interclass sensitivities may depend on the values of the parameters or variables with respect to which calculations are made, average sensitivities are calculated from variable and parameter ranges defined by the user in the model database. The results of this analysis are stored in one square matrix like the one exemplified in Table III – class A, with nA variables, influences class z, with nz variables: imposed changes on variable V1A affect variable V2z by 0.3%; imposed changes on variable V2A affect variables V1z, V2z and Vnz by –0.1%, 0.05% and 0.0056%, respectively.

**Table III-** Matrix synthesizing inter-class sensitivity coefficients between different variables

|  |  | Class A | | | | Class z | | |
|---|---|---|---|---|---|---|---|---|
|  |  | Var V1A | Var V2A | ... | Var VnA | Var V1z | Var V2z | ... | Var Vnz |
| Class A | Var V1A | - | - |  | - | 0 | 0.003 |  | 0 |
| | Var V2A | - | - |  | - | -0.001 | 0.0005 |  | 6E-05 |
| | ... |  |  |  |  |  |  |  |  |
| | Var VnA | - | - |  | - | 0 | 0 |  | 0.0008 |
| ... |  |  |  |  |  |  |  |  |  |
| Class z | Var V1z | -0.0001 | 8E-05 |  | 0 | - | - |  | - |
| | Var V2z | 0.0035 | 6E-06 |  | -4E-05 | - | - |  | - |
| | ... |  |  |  |  |  |  |  |  |
| | Var Vnz | 0.0006 | -3E-05 |  | 0.0001 | - | - |  | - |

After the Calibration Agent acquired this basic knowledge about the simulated system, it starts the calibration process itself:

(i) It selects the variable (Y) with the worst fit to observed data.
(ii) Secondly, it selects the parameter (P) or variable (Z) of its own class or of another class to which it is more sensitive. If it is a variable (Z), it selects the parameter (Pz) to which this influencing variable is more sensitive.
(iii) It changes the value of P or Pz in order to increase/decrease the average value of the variable under calibration (Y) towards the desired direction directly or indirectly through Z.
(iv) It runs the model.
(v) At the end of the simulation, it measures model lack of fit, adequacy and reliability, considering the overall model results.
(vi) It proceeds to (iii) iteratively until the desired range of the influencing parameter has been completely covered.
(vii) It chooses the best value of P or Pz.
(viii) It proceeds to the next variable with the worst fit and restarts the process from (ii).

After the first round across all model variables under calibration, the process may be repeated until the desired fit is obtained, model improvement stabilized or the calibration is stopped by the user.

The calibration system architecture with the Calibration Agent, EcoDyn application and data (observed data and model database) is shown in Figure 6. The user manages the agent actions and the EcoDyn activity and can manipulate the data present in the system, as the calibration process proceeds.

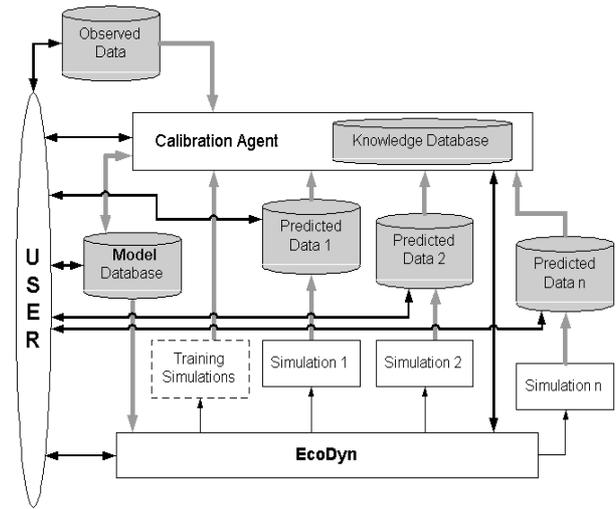

**Figure 6-** Calibration System architecture

## V. CONCLUSIONS AND FUTURE WORK

Agent-based ecological model calibration is a new and very promising approach enabling to fully automate a very complex and tedious problem to solve manually.

Controlled experiences have been made to test the validity of this new approach for model calibration. More intensive tests are being done with ecological models found in [7][12][13].

The result of this work will be applied in the calibration of the Ria Formosa (Algarve) model, in the DITTY project [10].

## VI. ACKNOWLEDGEMENTS

This work was supported by the DITTY project - "Development of an Information Technology Tool for the Management of European Coastal Lagoons under the influence of river-basin runoff", contract number EVK3-2001-00226, EU.

## REFERENCES

[1] Scholten H. & M.W.M. Van der Tol, Quantitative validation of deterministic models: when is a model acceptable? The proceedings of the Summer Computer Simulation Conference, SCS, San Diego, CA, USA: 404-409 (ISBN # 1-56555-149-4) , 1998.

[2] Mesplé, F., Troussellier, M., Casellas, C. & P. Legendre, Evaluation of simple statistical criteria to quality a simulation, Ecological Modelling 88 (1996): 9-18, 1995.

[3] Scholten H., Van der Tol, M.W.M. & A.C. Smaal, Models or measurements? Quantitative validation of an ecophysiological model of mussel growth and reproduction. Paper presented at the ICES Annual Science Conference, Cascais, Portugal, 1998.

[4] Janssen, P.H.M. & P.S.C. Heuberger, Calibration of process-oriented model, Ecological Modelling 83 (1995): 55-66, 1993.

[5] Norvig, P. & S.J. Russel, Artificial Intelligence: a modern approach, Englewood Cliffs, Prentice Hall, 1995.

[6] Weiss, G., editor, Multi-Agent Systems: A Modern Approach to Distributed Artificial Intelligence, MIT Press, 1999.

[7] Wooldridge, M., An Introduction to Multi-Agent Systems, John Wiley & Sons, Ltd, 2002.

[8] Reis, L.P., Coordination in Multi-Agent Systems: University Management and RoboCup Applications, PhD Thesis, Faculty of Engineering of the University of Porto, 2003 (in Portuguese).

[9] Jørgensen, S.E. & G. Bendoricchio, Fundamentals of Ecological Modelling, Elsevier Science Ltd, 3rd edition, 2001.

[10] The DITTY project description [online]. Available at http://www.dittyproject.org [visited March, 25 2004]

[11] Bacher, C., Duarte, P., Ferreira, J.G., Héral, M. & O. Raillard, Assessment and comparison of the Marennes-Oléron Bay (France) and Carlingford Lough (Ireland) Carrying Capacity with ecosystem models. Aquatic Ecology 31 (4): 379 - 394, 1998.

[12] Hawkins, A. J. S., Duarte, P., Fang, J. G., Pascoe, P. L., Zhang, J. H., Zhang, X. L. & M. Zhu., A functional simulation of responsive filter-feeding and growth in bivalve shellfish, configured and validated for the scallop Chlamys farreri during culture in China. Journal of Experimental Marine Biology and Ecology 281: 13-40, 2002.

[13] Duarte, P., Meneses, R., Hawkins, A.J.S., Zhu, M., Fang, J. & J. Grant, Mathematical modelling to assess the carrying capacity for multi-species culture within coastal waters, Ecological Modelling 168 (2003): 109-143, 2003.

[14] Vreugdenhil, C.B., Computatinal hydraulics, An introduction, Springer-Verlag, 1989.

[15] Knauss, J.A., Introduction to Physical Oceanography, Prentice-Hall, Englewood Cliffs, NJ, 1997 (referred in [13]).

[16] Neves, R.J.J.,Ètude expérimentale et modélisation mathématique des circulations transitoire et residuelle dans l'éstuaire du Sado, Ph.D thesis, Université de Liège, 1985 (referred in [13]).

[17] Odum, H.T., System ecology: An introduction, Wiley, Toronto, 1983 (referred in [13]).